\documentclass[11pt]{article}

\usepackage[top=1in, bottom=1in, left=1in, right=1in]{geometry}

\makeatletter
\newcommand*{\rom}[1]{\expandafter\@slowromancap\romannumeral #1@}
\makeatother

\newcommand{\<}{\langle}
\renewcommand{\>}{\rangle}

\newcommand{\beq}{\begin{equation}}
\newcommand{\eeq}{\end{equation}}

\newcommand{\E}{\mathbb{E}}

\renewcommand{\P}{\mathbb{P}}

\newcommand{\br}{\mathrm{\bf r}}

\newcommand{\bx}{\mathrm{\bf x}}

\newcommand{\bz}{\mathrm{\bf z}}

\newcommand{\bb}{\text{\boldmath $b$}}

\newcommand{\bbeta}{\text{\boldmath $\beta$}}

\newcommand{\btheta}{\text{\boldmath $\theta$}}

\newcommand{\bSigma}{\text{\boldmath $\Sigma$}}

\def\id{{\boldsymbol I}}

\newcommand{\sB}{\mathsf{B}}

\def\eps{\varepsilon}

\def\sV{{\sf V}}
\def\sT{{\sf T}}

\newcommand{\hf}{\widehat \bff}

\usepackage{mathrsfs}
\usepackage{caption}
\usepackage{subcaption}
\usepackage{balance}
\usepackage{hyperref}
\usepackage{graphicx}
\usepackage{booktabs}
\usepackage{xcolor}
\usepackage{amsmath}
\usepackage{amsfonts}
\usepackage{amsthm}
\usepackage[showdeletions]{color-edits}
\usepackage{float} 
\usepackage{amsmath}
\usepackage{color}
\usepackage{mathtools}
\usepackage{graphicx} 
\usepackage[LGR,T1]{fontenc}
\usepackage{mathrsfs}
\hypersetup{
    colorlinks,
    linkcolor={blue!80!black},
    citecolor={green!50!black},
}

 \usepackage[mathscr]{euscript}
 \DeclareSymbolFont{rsfs}{U}{rsfs}{m}{n}
 \DeclareSymbolFontAlphabet{\mathscrsfs}{rsfs}

\newtheorem{assumption}{Assumption}

\usepackage[utf8]{inputenc}

\def\<{\langle}
\def\>{\rangle}

\def\Err{\mathsf{Err}_{\mbox{\tiny\sf test}}}
\def\EBayes{\mathsf{Err}_{\mbox{\tiny\sf Bayes}}}

\def\bbeta{{\boldsymbol \beta}}
\def\bb{{\boldsymbol b}}

\def\bx{{\boldsymbol x}}

\def\bfzero{{\boldsymbol 0}}
\def\id{{\boldsymbol I}}

\def\bk{{\boldsymbol k}}
\def\bz{{\boldsymbol z}}

\def\bx{{\boldsymbol x}}

\def\bSigma{{\boldsymbol \Sigma}}
\def\btheta{{\boldsymbol \theta}}

\def\hbtheta{\hat{\boldsymbol \theta}}

\def\stest{\mbox{\tiny\rm test}}

\def\E{{\mathbb E}}
\def\P{{\mathbb P}}

\def\Loss{\Delta}

\def\hy{\hat{y}}

\def\sT{{\sf T}}

\def\eps{{\varepsilon}}

\def\hf{\hat{f}}

\def\br{{\boldsymbol r}}
\def\reals{\mathbb R}

\def\cuB{\mathscrsfs{B}}
\def\cuV{\mathscrsfs{V}}

\def\sD{{\sf D}}
\def\sV{{\sf V}}
\def\sB{{\sf B}}

\def\sd{{\sf d}}
\def\bSigma{{\boldsymbol\Sigma}}
\newtheorem{theorem}{Theorem}
\newtheorem{lemma}{Lemma}

\begin{document}

\title{Scaling Training Data with Lossy Image Compression}
\author{
Katherine L. Mentzer \;\; and \;\;
Andrea Montanari\thanks{Granica}}
\date{\today}
\maketitle

\begin{abstract}
  Empirically-determined scaling laws have been broadly successful in predicting 
  the evolution of large machine learning models with training data and number
  of parameters. As a consequence, they have been useful for optimizing the allocation
  of limited resources, most notably compute time. 

  In certain applications, storage space is an important constraint, and data 
  format needs to be chosen carefully as a consequence. Computer vision is a prominent example:
  images are inherently analog, but are always stored in a digital format using a finite number of bits.
  Given a dataset of digital images, the number of bits $L$ to store each of them can be further reduced 
  using lossy data compression. This, however, can degrade the quality of the model trained 
  on such images, since each example has lower resolution.

  In order to capture this trade-off and optimize storage of training data, 
  we propose a `storage scaling law' that describes the joint
  evolution of test error with sample size and \emph{number of bits per image}.
  We prove that this law holds within a stylized model for image compression,
  and verify it empirically  on two computer vision tasks, extracting the relevant parameters.
  We then show that this law  can be used to optimize the lossy compression level. 
  At given storage, models trained on optimally compressed images present a significantly smaller test error 
  with respect to models trained on the original data. Finally, we investigate the potential benefits
  of randomizing the compression level.
\end{abstract}

\section{Introduction}

\subsection{Background and motivation}

Training in machine learning has reached a scale at which storage is becoming an important concern. The storage size of a recent published research datasets ---LAION-5B
\cite{schuhmann2022laion}--- is of the order of 100 TB, and this number is
significantly larger in industry. Most companies in data-intensive domains
(e.g. recommendation systems or autonomous vehicles) train models on petabyte scale data. Neural scaling laws \cite{kaplan2020scaling,hoffmann2022training}
promise that any increase in the number of training samples will produce a predictable reduction\footnote{This of course holds under the condition
that the model complexity is scaled simultaneously.} in the error rate thus
providing a strong incentive towards accumulating increasingly massive datasets. 

This trends suggest a natural question: 
\begin{quote}
How should data scaling be extended under storage constraints?
\end{quote}
Part of the answer is suggested by the observation that a significant fraction of these data is, at a fundamental level, analog (images, audio, video). 
Therefore, any digital format captures the original signal only with a finite degree 
of resolution. The resolution can be changed, therefore changing the required storage space.
The `right' level or resolution must ultimately be decided  
on the basis of a utility criterion.

Throughout this paper we will focus ---to be definite--- on the case of 
images and machine learning tasks in computer vision. 
Prevalent image formats, such as PNG or JPEG, target human perceptual metrics.
These formats have lossy modes that allow to reduce the memory
footprint at the cost of some image quality degradation. Nevertheless, the common practice is to use the highest available quality level or, at worst, to
keep a format that is `visually lossless.' This approach
is not well justified and potentially wasteful, if data are used for machine learning training. Using more aggressive compression levels has the potential of addressing the storage constraint.

Several questions arise naturally:
\begin{enumerate}
\item How do models trained on (lossily) compressed images compare
to models trained on uncompressed ones?
\item What is the optimal compression level to be used under a given storage 
constraint?
\item How does does lossy compression affect data scaling laws?
\end{enumerate}
While there has been some work on these questions in the past 
\cite{dziugaite2016study, das2017keeping, yang2021compression,dodge2016understanding},
even the most basic one is not settled. In this paper we instead hypotesize a complete picture
and show that it can be used to push data scaling in a more efficient way.

\subsection{Summary of results}

We contend that, when storage is an important constraint, we should 
think about scaling laws not in terms of number of samples but (also) in terms
of bits. Our main finding is that machine learning generalization 
can be described, in a suitable storage-limited regime, by a bits-samples scaling law.

Consider a setting in which we are given $n$ training samples
$\{\bz_i\}_{i\le n}$. To be definite, in an image classification task,
$\bz_i=(\bx_i,y_i)$, where $\bx_i$ is an image and $y_i$ is the corresponding label (for a $K$-classes task, $y_i\in\{1,\dots,K\}$). We think 
of $\bz_i=(\bx_i,y_i)$ as an object potentially requiring infinitely many bits to be specified. For instance $\bx_i$ could be an analog grayscale image (modeled as a function  $\bx_i:[0,1]^2\to \reals$).

We train a machine learning model using an $L$-bits representation 
of the above samples, denoted by $\bz_i(L)$. For instance, in the case of images, we could use JPEG or JXL lossy modes to obtain $L$-bits representations. Letting $\hf(\,\cdot\, ;n,L)$ denote the model 
trained on such $L$-bits images, we denote by $\Err(n,L)$ be the test error
on original (uncompressed images):
\begin{align}
\Err(n,L) = \E\, \Loss\big(y_i, f(\bx;n,L)\big)\, .
\end{align}
Here  $\Loss(y, \hy)$ is a suitable loss function for the task of interest. 
For instance this could be classification loss $\Loss(y, \hy)=1$ if $\hy\neq y$
and $\Loss(y, \hy)=0$ if $\hy= y$.

We hypothesize that, in a suitable regime of $n,L$, the test error is well approximated
by the following scaling law
\begin{align}
\Err(n,L) \approx \Err^{*} + A\cdot n^{-\alpha}+B\cdot L^{-\beta}\, ,\label{eq:FirstScaling}
\end{align}
for some constants $\Err^{*}$, $A$, $B$ and scaling exponents $\alpha$, $\beta>0$.
It is important to emphasize that these parameters will depend on the specific
setting. Namely, they might depend on the machine learning task, on the dataset, 
on the model, on the compression algorithm, and so on. In particular, $\Err^*$ should not be interpreted
as the Bayes error,. but only as the minimal error attainable by the model under consideration.
We discuss these and other limitations and generalizations of 
Eq.~\eqref{eq:FirstScaling} in Section \ref{sec:Discussion}.

When verified, Eq.~\eqref{eq:FirstScaling} provides a principled way to
chose the level of compression for training data, by solving storage-constrained optimization problem
(here, we denote by $s$ the total available storage):
\begin{equation}\label{eq:opt-scaling}
\begin{aligned}
\mbox{minimize}&\;\;\; A\cdot n^{-\alpha}+B\cdot L^{-\beta}\, ,\\
\mbox{subj. to}&\;\;\; n\cdot L=s\, .
\end{aligned}
\end{equation}
\sloppy This yields the optimal solution $L_*(s) = C s^{\alpha/(\alpha+\beta)}$,  $n_*(s) = C^{-1} s^{\beta/(\alpha+\beta)}$,
with $C=(\beta B/\alpha A)^{1/(\alpha+\beta)}$
with corresponding test error
\begin{align}
\Err\big(n_*(s),L_*(s)\big) \approx \Err^{*} + C_E\cdot s^{-\nu}\, ,\;\;\;\nu = \frac{\alpha\beta}{\alpha+\beta}\, ,
\label{eq:OptScaling}
\end{align}
for a constant $C_E$.
Interestingly, the resulting storage scaling exponent is not given by either the sample size 
exponent $\alpha$ nor by the bits exponent $\beta$, but by their harmonic mean.

The rest of this paper is devoted to providing evidence for the the scaling
law \eqref{eq:FirstScaling} and its utility in optimally scaling 
the sample size $n$ and the bits per samples $L$ (hence the compression level):
\begin{enumerate}
    \item In Section \ref{sec:Experiments} we present results of numerical experiments on three computer vision tasks: image classification,  semantic segmentation, and object detection. 
    We verify that the test error is well described by the scaling law \eqref{eq:FirstScaling}, and that optimizing the predicted
    tradeoff between bits and samples yields near-optimal scaling of test error with storage.\footnote{The code is available at \url{https://github.com/granica-ai/LossyCompressionScalingKDD2024}.}
    \item In Section \ref{sec:Model}, we propose a very simple stylized model for image compression. We prove that the scaling  law \eqref{eq:FirstScaling}
    naturally arise as a consequence of the multiresolution structure of natural images.
\end{enumerate}
Finally, we discuss limitations and extensions of the present approach in Section \ref{sec:Discussion}.

\subsection{Related Work}

Scaling laws have proven a useful tool to capture the evolution 
of the test error with model complexity and number of training samples.
In their most successful version \cite{hestness2017deep, rosenfeld2019constructive}, 
they describe the dependence of $\Err$ on the number of training samples $n$, 
and the number of model parameters $m$, while keeping fixed the 
model architecture, and the specifics of the training algorithm. 
A scaling law postulates a power-law dependence on these two resources:
$\Err(n,m)\approx a\cdot n^{-\alpha}+b\cdot m^{-\gamma}$.
In this paper, we follow the same line of work and propose to treat the
number of bits per sample as the relevant resource (in addition
to the number of samples), in the storage-limited regime.

Motivated by the practical utility of scaling laws, a number of authors
have studied generalizations to incorporate the dependence on computational 
resources \cite{henighan2020scaling, kaplan2020scaling, bahri2021explaining, 
alabdulmohsin2022revisiting}, their applicability to finetuning a model that has been 
pretrained on a large dataset \cite{hernandez2021scaling, tay2021scale}, their 
extrapolation properties \cite{hoffmann2022training}, and the impact of data 
repetition in the data-constrained regime \cite{muennighoff2023scaling}.
Such laws have found their most important application in the
allocation of resource to data scaling versus compute scaling.

In practice, images are stored using a number of different codecs, with JPEG \cite{jpeg} and 
PNG \cite{png} among the most popular lossy and lossless methods, respectively. More modern codecs, 
such as JPEG XL \cite{alakuijala2019jpeg}, take advantage of human perception to achieve higher compression levels for approximately the same amount of visual or perceptual change to the image. Neural image compression is also an active area of research, with findings indicating that the human perception of an image can be largely preserved with many fewer bits \cite{rombach2022high, theis2022lossy}.

Lossy compression of training images  is implicit in the widespread use
of JPEG. However, the consequences of compression and
image resolution/loss on the resulting models are poorly understood, and the commonly accepted
practice is to store training images at the highest available resolutions
(`lossless' or `visually lossless').
Nonetheless, when images are lossily compressed, the impact on training is not well understood.
Several studies suggest that lossy image compression may help computer vision models by removing distracting noise or reducing adversarial perturbations \cite{dziugaite2016study, das2017keeping, yang2021compression}. Other
studies point out that lossy compression also discards information in the image, reducing image quality and sometimes leading to less accurate predictions \cite{dodge2016understanding}.


\section{Empirical Results}
\label{sec:Experiments}

\subsection{Set up}
In this section, we investigate the impact of scaling training data storage for three computer vision tasks:
$(1)$~An image classification task using the multi-class Food101 dataset \cite{bossard14} and a ResNet50 architecture \cite{he2016deep} pretrained on ImageNet-1K \cite{pytorch_weights}; $(2)$~a semantic segmentation task using the Cityscapes dataset \cite{Cordts2016Cityscapes} and a b5 SegFormer architecture \cite{xie2021segformer}, also pretrained on ImageNet-1K; $(3)$~a object detection task on aerial images using the iSAID dataset \cite{waqas2019isaid} and a a Mask R-CNN architecture \cite{He_2017}.

As anticipated in the introduction, we are interested in 
the dependence of test error on $n$ (the number of of training images)
and $L$ (which we redefine here as number of bytes per image). 
We achieve different values of $n$ by selecting subsets of the training data. We achieve different values of $L$ by using different levels of JPEG XL (JXL) lossy image compression \cite{alakuijala2019jpeg}. Note that a fixed byte size
is difficult to achieve, so we use $L$ do denote the average byte size instead.

The JXL compression level is determined by a perceptual distance metric, Butteraugli distance, with a higher distance corresponding to stronger compression and lower image quality. We train models on a grid of $n$ and $L$ values. Test error, $\Err$,
is measured in percent of images incorrectly classified for the classification task and one minus the mean intersection over union (mIoU) metric for the semantic segmentation task. For the object detection task, we select the bounding box mean average precision on small objects (less than $32^2$ pixels)---this metric is one of the standard COCO evaluation metrics \cite{lin2014microsoft}, and we select it to explore the impact of compression on smaller features in images. 

For each set of data gathered in this way, we fit the scaling
law \eqref{eq:FirstScaling} to the measured test error, to extract parameters
$\Err^{*}$, $A$, $B$, and exponents $\alpha$, $\beta$.

\subsection{Scaling curves}

\begin{figure*}[h]
    \centering
    \begin{subfigure}[t]{0.32\textwidth}
        \centering
        \setcounter{subfigure}{0}
        \includegraphics[width=\textwidth]{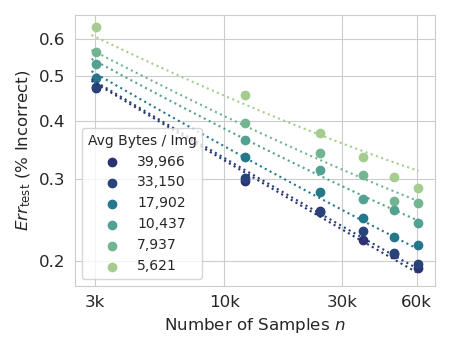}
        \caption{Classification task misclassification error scaling in $n$}
        \label{fig:food101-scaling-n}
    \end{subfigure}
    \begin{subfigure}[t]{0.32\textwidth}
        \centering
        \setcounter{subfigure}{2}
        \includegraphics[width=\textwidth]{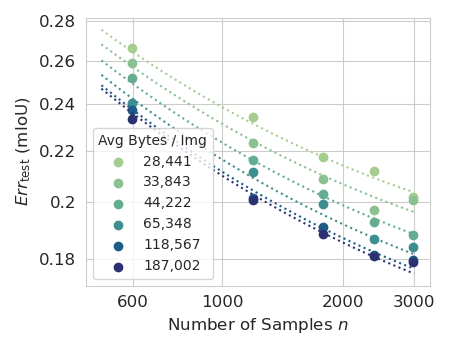}
        \caption{Semantic segmentation mean IoU scaling in $n$}
        \label{fig:cityscapes-scaling-n}
    \end{subfigure}
    \begin{subfigure}[t]{0.32\textwidth}
        \centering
        \setcounter{subfigure}{4}
        \includegraphics[width=\textwidth]{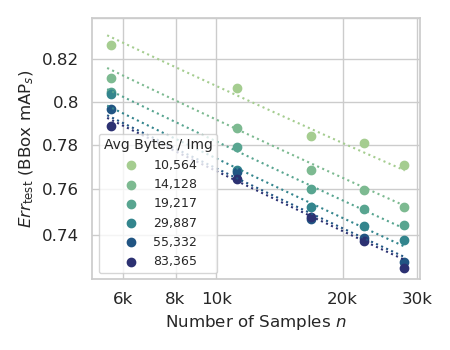}
        \caption{Object detection mean accuracy scaling in $n$}
        \label{fig:isaid-scaling-n}
    \end{subfigure}

    \begin{subfigure}[b]{0.32\textwidth}
        \centering
        \setcounter{subfigure}{1}
        \includegraphics[width=\textwidth]{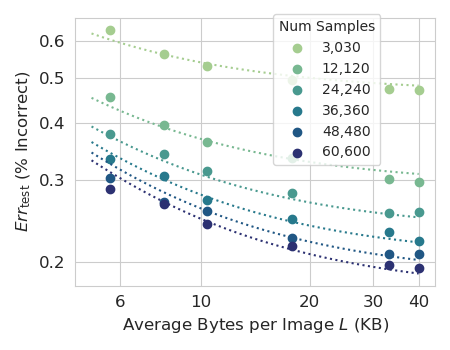}
        \caption{Classification task misclassification error scaling in $L$}
        \label{fig:food101-scaling-l}
    \end{subfigure}
    \begin{subfigure}[b]{0.32\textwidth}
        \centering
        \setcounter{subfigure}{3}
        \includegraphics[width=\textwidth]{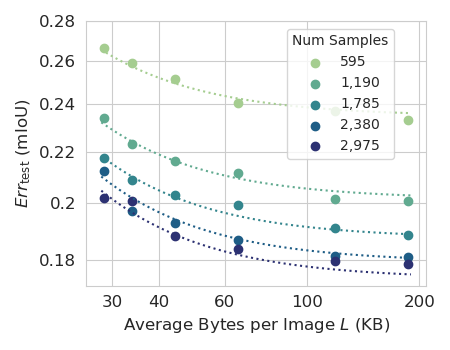}
        \caption{Semantic segmentation mean IoU scaling in $L$}
        \label{fig:cityscapes-scaling-l}
    \end{subfigure}
    \begin{subfigure}[b]{0.32\textwidth}
        \centering
        \setcounter{subfigure}{5}
        \includegraphics[width=\textwidth]{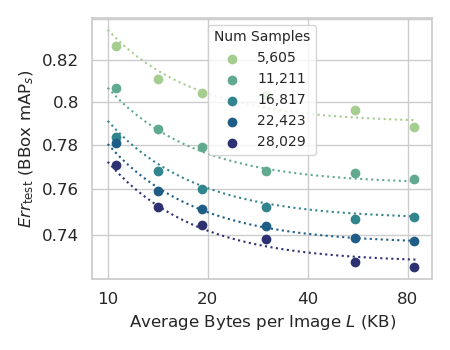}
        \caption{Object detection mean accuracy scaling in $L$}
        \label{fig:isaid-scaling-l}
    \end{subfigure}
    \caption{Test error scaling for the image classification, semantic segmentation, and object detection tasks. Circles represent empirical results, and the dotted line indicates the scaling curve fit to these observations.}
    \label{fig:scaling}

\end{figure*}

For each task, we select $n$ and $L$ values in the following way. The $n$ values for each dataset are approximately 20\%, 40\%, 60\%, 80\%, and 100\% of a train split for each dataset. In the case of the image classification task, these samples are stratified, maintaining class balance in each subset. The range of $n$ values varies significantly between tasks due to the differing sizes of the datasets. To achieve different $L$ values, we compress images using JXL to Butteraugli distances 
${\rm Bd}\in \{1, 2, 4, 7, 10, 15\}$, where higher distances 
correspond to smaller average byte sizes $L$. Training procedures mimic the training method for each model's original paper, and the approaches are detailed in \autoref{appendix:training}.

\autoref{fig:scaling} shows the resulting scaling curves for the three computer vision tasks. Circles indicate empirical model test errors, and the dotted line illustrates curve that we fit to these average test errors. \autoref{tab:curve-params} shows the parameters for these fit curves. 

\begin{table}[h]
    \centering
    \begin{tabular}{ccccccc}
    \toprule
        \textbf{Task} && $A$ & $B$ & $\Err^{*}$ & $\alpha$ & $\beta$ \\
        \cmidrule{1-1} \cmidrule{3-7}
        Classification && $6.7$ & $1.4 \times 10^3$ & $0.0$ & $0.33$ & $ 1.06$ \\
        Segmentation && $4.3$ & $4.3 \times 10^5$ & $0.1$ & $0.59$ & $1.6$ \\
        Detection && $1.0$ & $2.7 \times 10^5$ & $0.3$ & $0.09$ & $1.7$ \\
        \bottomrule
    \end{tabular}
    \caption{Scaling curve fit parameters for each task, where curves take the form $\Err(n,L) \approx \Err^{*} + A\cdot n^{-\alpha}+B\cdot L^{-\beta}$.}
    \label{tab:curve-params}
\end{table}

We observe that the test error decreases with $n$ and $L$
as expected, and that this behavior is well approximated by the scaling 
law \eqref{eq:FirstScaling}. We note that while the plots are shown with log-scale axes, the curves are not precisely linear. This is a consequence of the fact that the scaling law \eqref{eq:FirstScaling} is
the sum of contributions. If we vary only $n$ or only $L$, the  excess error $\Err(n,L)-\Err^*$
will saturate at a strictly positive level.
We also observe that the exponent $\beta$ is typically larger than $\alpha$,
indicating that the effect of short-scale details decays rather rapidly. 
However, the last conclusion is likely to be sensitive to the specific task under study. Furthermore, this power law pattern is not unique to these chosen metrics; we find other measures of error follow similar patterns, shown in \autoref{appendix:additional-metrics}. 

\subsection{Optimizing $n$ and $L$}

Given a scaling law that fits the data, and a fixed storage value $s$, we can
determine the optimal $n_*(s)$ and $L_*(s)$ to minimize the predicted 
test error, as per Eq.~\eqref{eq:opt-scaling}. 
The behavior $n_*(s)$ and $L_*(s)$ according to our model suggests that, in the storage-limited regime, opting for more images of lower quality can improve model performance over fewer, less compressed images.

\begin{figure*}
    \centering
    \begin{subfigure}[b]{0.32\textwidth}
        \centering
        \setcounter{subfigure}{0}
        \includegraphics[width=\textwidth]{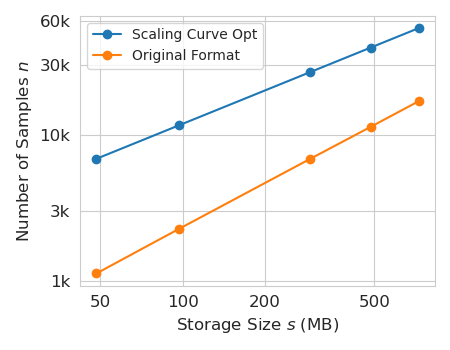}
        \caption{Classification, number of samples}
        \label{fig:food101-s-vs-n}
    \end{subfigure}
    \begin{subfigure}[b]{0.32\textwidth}
        \centering
        \setcounter{subfigure}{2}
        \includegraphics[width=\textwidth]{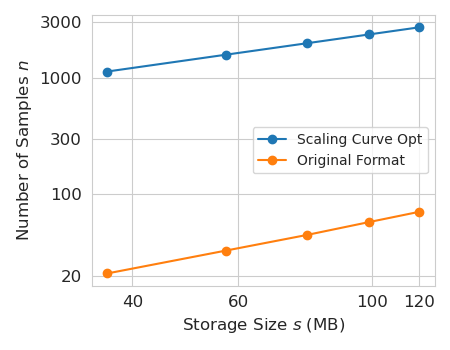}
        \caption{Segmentation, number of samples}
        \label{fig:cityscapes-s-vs-n}
    \end{subfigure}
        \begin{subfigure}[b]{0.32\textwidth}
        \centering
        \setcounter{subfigure}{4}
        \includegraphics[width=\textwidth]{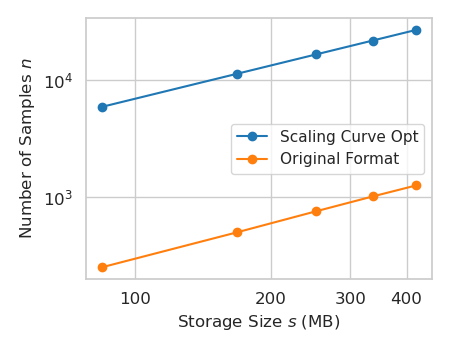}
        \caption{Detection, number of samples}
        \label{fig:isaid-s-vs-n}
    \end{subfigure}
    
    \begin{subfigure}[b]{0.32\textwidth}
        \centering
        \setcounter{subfigure}{1}
        \includegraphics[width=\textwidth]{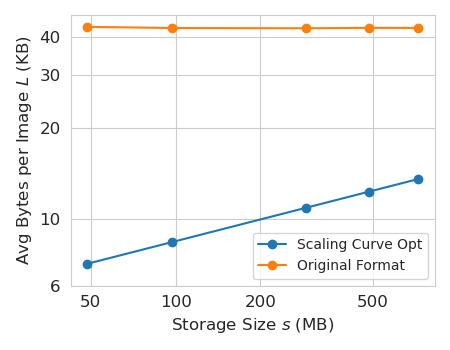}
        \caption{Classification, number of bytes}
        \label{fig:food101-s-vs-l}
    \end{subfigure}
    \begin{subfigure}[b]{0.32\textwidth}
        \centering
        \setcounter{subfigure}{3}
        \includegraphics[width=\textwidth]{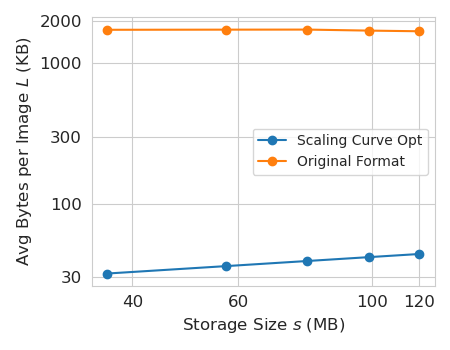}
        \caption{Segmentation, number of bytes}
        \label{fig:cityscapes-s-vs-l}
    \end{subfigure}
    \begin{subfigure}[b]{0.32\textwidth}
        \centering
        \setcounter{subfigure}{5}
        \includegraphics[width=\textwidth]{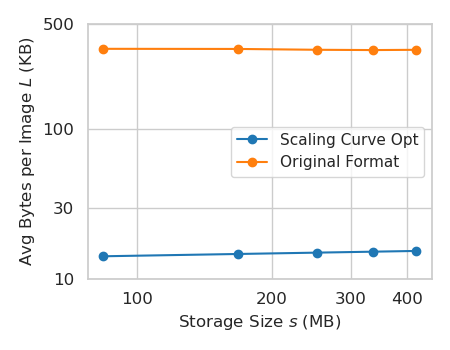}
        \caption{Detection, number of bytes}
        \label{fig:isaid-s-vs-l}
    \end{subfigure}
    \caption{Values of $n$ and $L$ as a function of total storage size for each task under the optimal and original format training data.}
    \label{fig:s-vs-nl}
\end{figure*}

\autoref{fig:s-vs-nl} illustrates the optimal $n_*(s)$ and $L_*(s)$ values under our model for each task.
We compare number of images at a given storage level $s$ to the number of images in original format
(i.e. without lossy compression) that fit into the given storage.
The curves for the image classification task in \autoref{fig:food101-s-vs-n}-\subref{fig:food101-s-vs-l} illustrate that leaving images in their original JPEG format results in a larger higher $L$ and lower $n$ than  optimal, if the goal is to minimize test error. If the images are previously stored as PNGs, as in the semantic segmentation task and the object detection tasks, this effect is even more pronounced, as illustrated in \autoref{fig:cityscapes-s-vs-n}-\subref{fig:cityscapes-s-vs-l} and \autoref{fig:isaid-s-vs-n}-\subref{fig:isaid-s-vs-l}. \autoref{tab:opt_n_l} also captures these results. 
Hence, giving up some image quality (via lossy compression) to acquire a greater number of example images is preferable in this storage-limited setting.



\begin{table*}[h]
    \centering
    \begin{tabular}{ccccccccccc}
    \toprule
        \multicolumn{3}{c}{\textbf{Classification Task}} && \multicolumn{3}{c}{\textbf{Segmentation Task}} && \multicolumn{3}{c}{\textbf{Detection Task}} \\
        $s$ & $n_*(s)$ & $L_*(s)$ && $s$ & $n_*(s)$ & $L_*(s)$ && $s$ & $n_*(s)$ & $L_*(s)$ \\\cmidrule{1-3} \cmidrule{5-7} \cmidrule{9-11}
        48,480,000  & 6,838  & 7,087    && 36,360,000    &1,137 & 31,965    && 84,087,000   & 5,913     & 14,219    \\
        96,960,000  & 11,588 & 8,363    && 57,267,000    &1,586 & 36,090    && 168,174,000  & 11,417    & 14,730    \\
        290,880,000 & 26,737 & 10,876   && 78,174,000    &1,991 & 39,254    && 252,261,000  & 16,776    & 15,037    \\
        484,800,000 & 39,441 & 12,289   && 99,080,999    &2,368 & 41,831    && 336,348,000  & 22,043    & 15,258    \\
        727,200,000 & 53,697 & 13,540   && 119,988,000   &2,725 & 44,016    && 420,435,000  & 27,243    & 15,433    \\
        \bottomrule
    \end{tabular}
    \caption{Optimal $n$ and $L$ for the classification, segmentation, and detection tasks, according to the fit scaling curves. $L$ is the average bytes per pixel; that is, the empirical dataset size in bytes divided by the number of images in the subset.}
    \label{tab:opt_n_l}
\end{table*}

\subsection{Test error for optimal $n$ and $L$}

The scaling curves suggest that the optimal trade-off between $n$ and $L$ requires more compression than used in standard image storage formats. 
We next explore the impact of optimizing this tradeoff on test error.
Namely, we consider the following strorage sizes 
$s \in [$4848, 9696, 29088, 48480, 72720$]\times 10^4$ (for classification)
$s \in [3636, 5726.7, 7817.4, 9908.1, 11998.8]\times 10^4$ (for segmentation), and $s \in [8408.7, 16817.4, 25226.1, 33634.8, 42043.5] \times 10^4$ (for detection) and train models
using the following data scaling schemes.

\begin{figure*}
    \centering
    \begin{subfigure}[b]{0.32\textwidth}
        \centering
        \includegraphics[width=\textwidth]{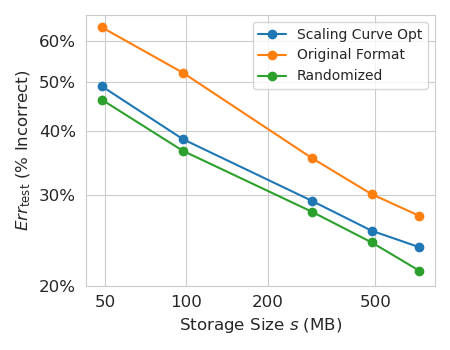}
        \caption{Image Classification}
        \label{fig:food101-s-curve}
    \end{subfigure}
    \begin{subfigure}[b]{0.32\textwidth}
        \centering
        \includegraphics[width=\textwidth]{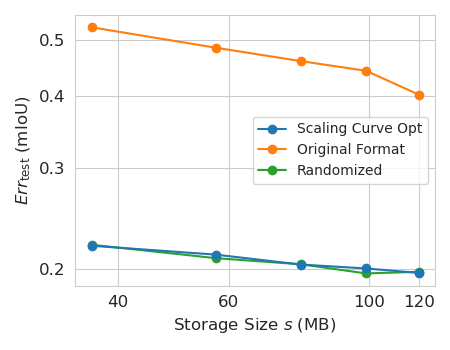}
        \caption{Semantic Segmentation}
        \label{fig:cityscapes-s-curve}
    \end{subfigure}
    \begin{subfigure}[b]{0.32\textwidth}
        \centering
        \includegraphics[width=\textwidth]{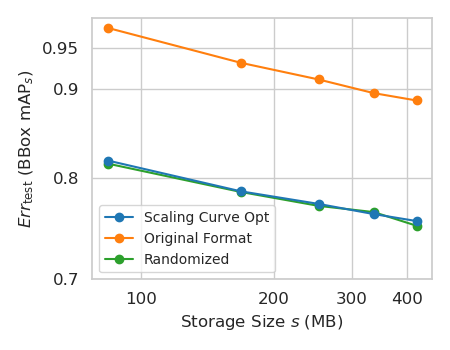}
        \caption{Object Detection}
        \label{fig:isaid-s-curve}
    \end{subfigure}
    \caption{Test error scaling in $s$ for each task task using
    different data scaling schemes.}
\end{figure*}


\paragraph{Scaling Optimal.} 
For each $s$ calculate the optimal $n_*(s)$ and $L_*(s)$ that minimize the 
test error according to the fitted scaling law.
We then select stratified subsets of the dataset of size $n_*(s)$ 
and search for the image compression level that achieves the raget 
$L_*(s)$ on average. Note that all images are compressed using the 
same compression parameters, but the bytes per image may vary between 
images based on image content and compression characteristics. We train a 
model using the same architecture and hyperparameters as the scaling curve 
models. To average over the effect of the specific subset selected 
for each model, we repeat this process 2 more times for a total of 
3 models trained at each $n_*(s)$ and $L_*(s)$ value. We report 
accuracy averaged over these 3 iterations. 

\paragraph{Original Format.} An alternative to allowing lossy compression is to 
retain full image resolution; to achieve the desired storage capacity, we simply 
discard images until the training images fit within the budgeted space. For each 
storage size $s$ used in the scaling optimal models, we train a model based on an 
original format subset that occupies the same amount of storage. The JXL image 
codec typically outperforms older image codecs such as PNG and JPEG; 
therefore, for a most fair comparison to the lossily compressed 
JXL images, we use the JXL version of the original data storage 
to calculate the storage size of a given subset—i.e., the Food101 
data, originally stored as JPEGs, is compressed using the JPEG to JXL 
transcoding that does not introduce further loss, and the Cityscapes and iSAID datasets,
originally stored as PNGs, are compressed using JXL lossless mode. In a slight
abuse of notation, we still refer to these JXL versions that do not introduce
further loss as `original format.' Under this encoding, the average bytes per pixel
is $L=$ 42,804 for the Food101 dataset, $L=$ 1,711,011 for the Cityscapes dataset, and $L=337,789$ for the iSAID dataset.
The resulting subsets are then used to train a model with the same parameters as the 
original scaling curves.

\paragraph{Randomized.} The scaling optimal models use the optimal $n_*(s)$ 
and $L_*(s)$ for a given fixed storage, where all images are compressed to the same 
level. We investigate the effect of heterogeneity in the compression levels.
One might argue that heterogeneous compression can lead to  models benefiting from exposure to both more and less compressed images.

Our randomization strategy is as follows. Using the same image subsets as the scaling  optimal training images, we compress each image to some level within 
a range of different compression levels. To generate the range of 
compression levels, the training images are randomly ordered. This 
ordering is then translated into a compression level, where the first 
ranked image is compressed to the lowest level, the last ranked image is compressed
to the highest level, and the remaining images are uniformly spaced in between. 
To determine the minimum and maximum compression level, we use the following 
procedure. The minimum compression level is the smallest non-negative JXL 
distance that leads to compression relative to the original storage format — in 
practice, a distance of 0.5 for the Food101 dataset and 0 for the Cityscapes and iSAID datasets. 
The maximum compression level is then increased until the subset reaches the desired 
compressed size or a distance of 15. If a distance of 15 is reached and the subset is 
still too large with regard to compressed size, the minimum compression level is 
increased until the images reach the desired compressed size. We selected the maximum 
distance of 15 due to diminishing reductions in file size beyond this distance. Once 
we find a mapping of images to compression levels that achieves the target storage 
size, we then train a model using the same parameters as the original scaling curves.

\autoref{fig:food101-s-curve} shows the test error of each of these strategies for 
the image classification task, averaged over three different realizations of model 
training. The original format models performs significantly worse than the scaling 
 optimal models. 
Furthermore, the randomized curve shows a small but consistent benefit over the 
uniformly compressed scaling curve optimal result, suggesting that this model indeed 
benefits from exposure to a variety of compression levels. 

\autoref{fig:cityscapes-s-curve} shows the same results for the segmentation task. 
Due to the Cityscapes images original PNG format, the number of losslessly compressed 
images that fit into the storage budget is very low (fewer than 100 samples). 
Consequently, models trained on so few images perform significantly worse than models 
trained on more, lossily-compressed images. Furthermore, the randomized results are 
noisier, with variable results compared to the uniformly compressed scaling curve 
optimal models. The object detection results, shown in \autoref{fig:isaid-s-curve}, 
follow a similar structure, as the original image format was also PNG.

\subsection{Comparison to naive compression}

\begin{figure}
    \centering
    \includegraphics[width=.45\textwidth]{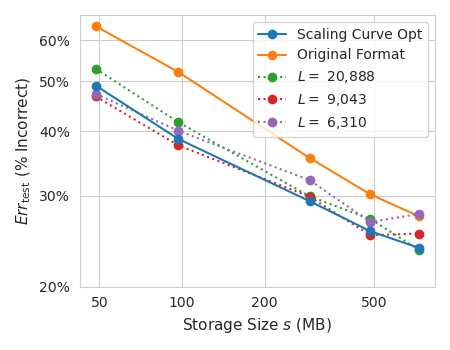}
    \caption{Test error for models trained on naively compressed data relative to test error for models trained on optimally compressed data and original format data.}
    \label{fig:naive-compression}
\end{figure}

We next explore the benefits of using scaling optimal compression over fixed-level
compression. \autoref{fig:naive-compression} compares the test error at the optimal 
values of $n$ and $L$ to the test error for other (fixed) compression levels. 
Similar to the original format curve in 
Figures~\ref{fig:food101-s-curve}~and~\ref{fig:cityscapes-s-curve}, 
we select the greatest number of images that fit within the storage budget at 
the given level of compression. For each value of $s$, we generate 3 different 
data subsets and train 3 models, reporting the average test error. 
reduce the variability due to the specific data subset, the
subsets are chosen to overlap as much as possible with the subsets used 
to train the 3 models that comprise the optimal curve. 

 We observe that fixed compression levels typically lead to the same or higher test error than the optimally selected value $L_*(s)$, though they
 have an advantage over original format data. At the smallest storage sizes, results are more variable---test error varies based on the specific subset of images used to train the model, leading to a less clear benefit from compression level 
 optimization. For example, at the lowest storage levels, the $L=$ 6,310 and $L=$ 9,043 models perform slightly better than the optimal $L=$ 6,838 model, despite the naive models landing on opposite sides of the $n$ versus $L$ trade-off relative to the optimal. This result can be attributed to the higher variability of model performance at lower storage sizes and the differences in the specific examples used for training.

\subsection{Test set compression}
The scaling law \eqref{eq:FirstScaling} predicts how models trained on differently compressed data performs on uncompressed test data. A natural question is how these models perform on compressed test sets. To investigate this question, we take the largest $n$ model trained at each compression level $L$ and evaluate it on a number of compressed test sets. The scaling models were trained on JXL images compressed to Butteraugli distances $\mathrm{Bd} \in \{1,2,4,7,10,15\}$, and we evaluate each model on a test set compressed to these same Butteraugli distances. The result is 6 different test accuracy results for each of 6 different models. 

\begin{figure*}[h]
    \centering
    \begin{subfigure}[b]{0.32\textwidth}
        \centering
        \includegraphics[width=\textwidth]{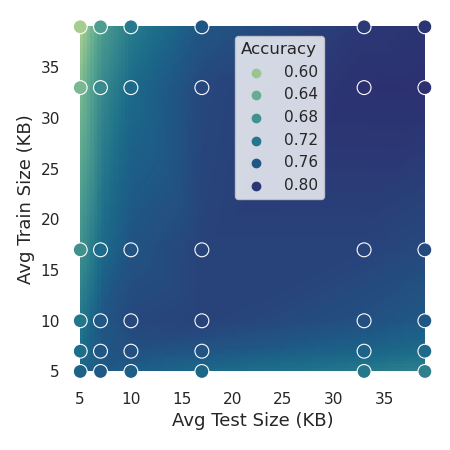}
        \caption{Image Classification}
        \label{fig:food101-test-compr}
    \end{subfigure}
    \begin{subfigure}[b]{0.32\textwidth}
        \centering
        \includegraphics[width=\textwidth]{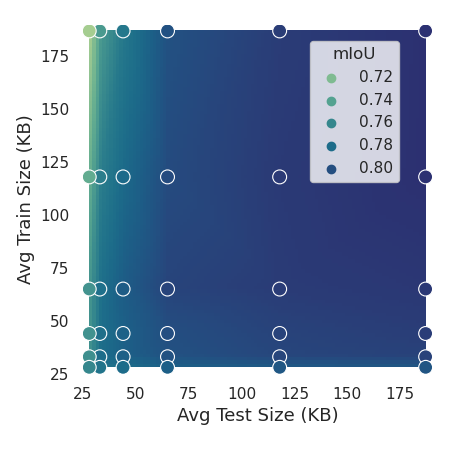}
        \caption{Semantic Segmentation}
        \label{fig:cityscapes-test-compr}
    \end{subfigure}
    \begin{subfigure}[b]{0.32\textwidth}
        \centering
        \includegraphics[width=\textwidth]{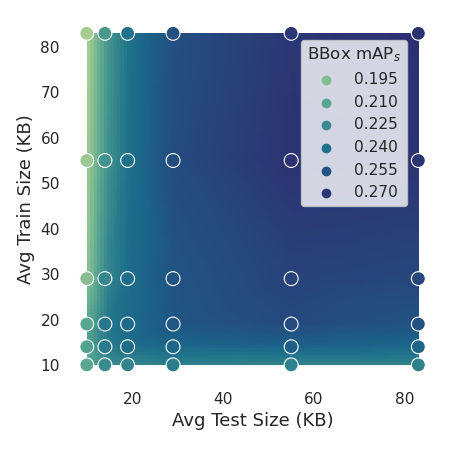}
        \caption{Object Detection}
        \label{fig:isaid-test-compr}
    \end{subfigure}
    \caption{Error as a function of both training data compression level and test data compression level for the classification, segmentation, and detection tasks. Circles indicate empirical results from model evaluation, and the background represents a linear interpolation of those points.}
    \label{fig:test-compr}
\end{figure*}

The circles in \autoref{fig:test-compr} represent these accuracies, and the background color is a linear interpolation of these observations. Unsurprisingly, the highest test accuracy is typically when the model is evaluated on data that is compressed to the same level as the training data (i.e., when the test and train data is most similar). More interestingly, for all three tasks, the largest reduction in performance is for models trained on low compression levels but evaluated on highly compressed data. Models trained on somewhat more compressed data see less performance degradation, demonstrating some robustness to variability in test data quality.
%
%
\section{A stylized model}
\label{sec:Model}

In this section we introduce a stylized model for machine learning from compressed images.
To motivate this model, consider the case of a grayscale image,
modeled as a function $f:[0,1]^2\to\reals$.
For image processing purposes, this is most conveniently represented in a multiscale
fashion \cite{daubechies1992ten,mallat1999wavelet}, 
e.g. via wavelet transform. Schematically, for $\br\in [0,1]^2$
\begin{align}
f(\br) = \sum_{\ell=0}^{\infty}\sum_{\bk\in [q_0^{\ell}]^2}
x^{(\ell)}_{\bk}\psi\big(q_0^{\ell}(\br-\bk q_0^{-\ell})\big)\, .
\end{align}
Here $\psi$ is the basis wavelet and $\{x^{(\ell)}_{\bk}:\,  \bk \in [q_0^{\ell}]^2\}$
are $q_0^{2\ell}$ coefficients at scale $\ell$. 

Part of the motivation for considering this representation stems from the fact that 
standard lossy compression schemes (such as JPEG or JXL) are fundamentally based on multiscale analysis, albeit in a simplified form \cite{donoho1998data}. To a first order,
lossy compression is achieved by quantizing to a coarser level (or completely setting to zero) smaller scale coefficients. 

Rather than noting explicitly  $\bk$ as a two-dimensional index, we will 
directly think an example as an infinite vector
\begin{align}
\bx = (\bx^{(0)}, \bx^{(1)}, \bx^{(2)}, \cdots)\, ,\;\;\; \bx^{(\ell)}\in \reals^{q^{\ell}}\, ,
\end{align}
with $\bx^{(\ell)} = (x^{(\ell)}_1,\dots,x^{(\ell)}_{q^{\ell}})$.
To make a connection with the image setting, we can choose $q=q_0^2$.
As a cartoon for a lossy compression algorithm, 
we consider the scheme that truncates the multiscale coefficients at level $m$
\begin{align}
\Phi_{m}(\bx) = (\bx^{(0)}, \bx^{(1)}, \bx^{(2)}, \cdots,\bx^{(m)})\, ,\;\;\;
L=\sum_{\ell=0}^{m} q^{\ell}\, .
\end{align}
We identify $L$ as the number of bits in our compressed representation. This does not change scaling laws, as long as each coordinate of $\bx$ can be represented by a constant (independent of $n,L$)
number of bits (one byte is common in image formats).

We consider the simplest model of supervised learning.
We are given samples $(y_i,\bx_i)\in \reals\times\reals^{\infty}$, $i\le n$, whereby
\begin{align}
y_i = \<\bx_i,\btheta\>+\eps_i \, ,\;\;\; \E(\eps_i^2)=\tau^2\, ,
\end{align}
where the noise variables $\eps_i$ have zero mean and are i.i.d. and independent of everything else.
Given such a model, it is natural to estimate the coefficients $\btheta\in\reals^{\infty}$
via ridge regression. This classical approach has recently drawn renewed interest
because of its mathematical equivalence to kernel ridge regression, and its connection
to neural networks in the linear regime.

Since we want to perform learning on compressed data, we use
the vectors $\Phi_{m}(\bx_i)$ in learning:
\begin{align}
\hbtheta_{\lambda}(n,L) = \arg\min_{\bb\in\reals^L}
\Big\{\sum_{i=1}^n\big(y_i-\<\bb,\Phi_m(\bx_i)\>\big)^2+\lambda\|\bb\|^2\Big\}\, .
\end{align}
We square loss also to define test error, namely
\begin{align}
\Err(n,L)=\E\Big\{\Big(y_{\stest}-\<\hbtheta_{\lambda}(n,L),\Phi_m(\bx_{\stest})\>\Big)^2\Big\}
\, ,\label{eq:StylizedTest}
\end{align}
where $(y_{\stest},\bx_{\stest})$ is a fresh sample distributed as the training data
$\{(y_i,\bx_i)\}_{i\le n}$. 

It is a well-documented empirical observation that small
scale components of natural images carry less energy than large scale components.
Further, energy typically decays as a power law of the length scale.
Since the length scale for block $\ell$ is $q_0^{-\ell}$, this translates into an 
exponential decay with $\ell$.
We assume a simple model that captures this behavior, namely the blocks $\bx^{(\ell)}$,
$\ell\ge 0$ are mutually independent with $\E\{\bx^{(\ell)}\} = \bfzero$ and
\begin{align}
 \E\{\bx^{(\ell)}(\bx^{(\ell)})^{\sT}\} = s_{\ell}\id_{q^{\ell}}\, ,
 \;\;\;\; K_1p^{-\ell}\le s_{\ell}\le K_2 p^{-\ell}\, ,
 \label{eq:CovAssumption}
\end{align}
where $p>1$ is a model parameter and $0< K_1< K_2$ are constants.
We will further assume the distribution of $\bx^{(\ell)}$ to satisfy some technical 
conditions stated in Assumption \ref{ass:X-assumption}, see Appendix \ref{app:Proofs}.
Note that the expected  total energy is $\E\|\bx\|^2\asymp \sum_{\ell\ge 0} (q/p)^{\ell}$:
in order for this to be finite, we require $q<p$.

Finally, it is natural to assume that smaller scale component will have decreasing 
importance for our supervised learning task. We model this by writing
the multiscale decomposition of the parameters $\btheta$ as
$\btheta = (\btheta^{(0)},\btheta^{(1)},\btheta^{(2)},\dots)$, and assuming,
for some $r<1$ and constants $0< \tilde K_1< \tilde K_2$:
\begin{align}
\tilde K_1 r^{\ell}\le  \|\btheta^{(\ell)}\|^2  \le \tilde K_2 r^{\ell}\, .\label{eq:BetaAssumption}
\end{align}

Under this model, the test error the test error \eqref{eq:StylizedTest} can be rewritten as
\begin{align}
\Err&(n,L) =\nonumber\\
&=\tau^2 + 
 \sum_{\ell=0}^m\E\{\<\hbtheta^{(\ell)}-\btheta^{(\ell)},\bx^{(\ell)}\>^2\}+
\sum_{\ell=m+1}^\infty\E\{\<\btheta^{(\ell)},\bx^{(\ell)}\>^2\}\nonumber\\
&=\tau^2 + \sum_{\ell=0}^m\E\{\<\hbtheta^{(\ell)}-\btheta^{(\ell)},\bx^{(\ell)}\>^2\}+\overline K_m\cdot (r/p)^m\, ,\label{eq:BasicDec}
\end{align}
where $K_1\tilde K_1r/(p-r)\le \overline K_m\le K_2\tilde K_2r/(p-r)$.
This expression corresponds to the decomposition of the test error
in three terms, reflecting the scaling form \eqref{eq:FirstScaling}: 
a fundamental limit $\tau^2$ (in the present case, this is just the Bayes error);
an statistical estimation error, corresponding to the sum
over $\ell\le m$; and a compression-induced error, corresponding to
the smalle length
scales, i.e. $\ell>m$.

The next result  shows that indeed this decomposition reproduces the 
scaling law \eqref{eq:FirstScaling}.
\begin{theorem}\label{thm:main}
Under the model described above, further assume $rp<1$, $q/p<1$ and
the conditions of Assumption \ref{ass:X-assumption} to hold.
Then there exist positive constants $0<A_1<A_2$, $0<B_1<B_2$, $0<C$, $0<c_0$
(depending on the constants $p,q,r,\tau$ in the assumptions),
such that, for all $n\le c_* L^{1+2\kappa}$, 
and $\lambda  \in [1/M,M]\cdot n^{(\kappa+1)/(2\kappa+1)}$ 
for some constant $M$ ($\kappa:=\log p /\log q$), 
the following holds with high probability
\begin{align*}
 \tau^2+A_1\cdot n^{-\alpha} +B_2L^{-\beta} \le \E_{\eps}\Err(n,L)\le \tau^2+A_2\cdot n^{-\alpha} +B_2L^{-\beta}\, ,
\end{align*}
with exponents
\begin{align}
\alpha =\frac{2\log p}{2\log p+\log q},\;\;\;\; \beta = \frac{\log(p/r)}{\log q}\, .
\end{align}
(In the above formulas, $\E_{\eps}$ denotes expectation with respect to the noise in the observations $\eps_1$,\dots, $\eps_n$.)
\end{theorem}
The proof of this result is given in Appendix \ref{app:Proofs},
and builds upon recent characterizations of high-dimensional ridge regression
\cite{cheng2022dimension,tsigler2023benign}.
We note that the same proof can be used also to characterize the behavior 
 $n> c_0L$. For the sake of brevity, 
 we omit to state that result here.

In fact our analysis provides a sharper (albeit more complicated) characterization 
of test error than what stated in Theorem \ref{thm:main}. In order to state this characterization, 
we introduce the following functions from \cite{cheng2022dimension}. First, define
$\lambda_*$ as the unique non-negative solution of
\begin{align}
n-\frac{\lambda}{\lambda_*} = \sum_{\ell=0}^{m}\frac{s_{\ell}q^{\ell}}{s_{\ell}+\lambda_*}\, .\label{eq:lambdastar}
\end{align}
We then define
\begin{align}
\sB(m,n)  &:= \frac{n\lambda_*^2}{n-\sD(m)}  \sum_{\ell=0}^{m}
\frac{s_{\ell}\|\btheta^{(\ell)}\|^{2}} {(s_{\ell}+\lambda_*)^2}\, ,\\
\sV(m,n) &:= \frac{\sD(m)}{n-\sD(m)}\, ,
\end{align}
where
\begin{align}
\sD(m)= \sum_{\ell=0}^{m}\frac{s^2_{\ell}q^{\ell}}{(s_{\ell}+\lambda_*)^2}\, .
\end{align}

\begin{lemma}\label{lemma:Ridge}
Under the assumptions of Theorem \ref{thm:main},
consider  $\lambda\ge n^{1-\kappa}/M$ for any constant $M$. Then we have
\begin{align*}
\E_{\eps}\Err(n,L) =\tau^2 + \cuB(m,n)+\tilde\tau^2_m \cuV(m,n)+
\overline K_m\cdot (r/p)^{m}\, .
\end{align*}
where (as above)  $K_1\tilde K_1r/(p-r)\le \overline K_m\le K_2\tilde K_2r/(p-r)$,
\begin{align}
    \tilde\tau^2_m  = \tau^2+\overline K_m\cdot (r/p)^{m}\, ,
\end{align}
and
\begin{align}
\cuB(m,n) &= \big(1+O(n^{-0.49})\big)\sB(m,n)\, ,\\
\cuV(m,n) &= \big(1+O(n^{-0.99})\big)\sV(m,n)\, .
\end{align}
 \end{lemma}

\begin{figure}
    \centering
    \begin{subfigure}[b]{0.45\textwidth}
        \centering
        \includegraphics[width=\textwidth]{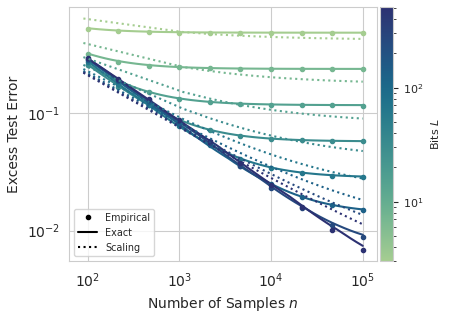}
        \caption{Excess test error scaling in $n$ for the stylized model.}
        \label{fig:stylized-n}
    \end{subfigure}
    \begin{subfigure}[b]{0.45\textwidth}
        \centering
        \includegraphics[width=\textwidth]{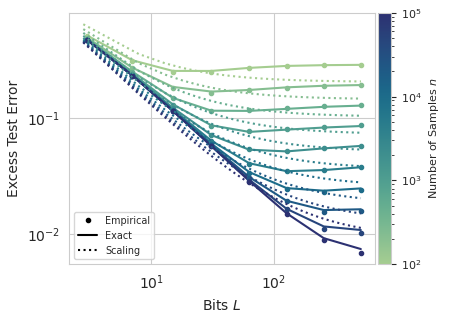}
        \caption{Excess test error scaling in $L$ for the stylized model.}
        \label{fig:stylized-L}
    \end{subfigure}
    \caption{Simulation results of the stylized model indicate similar scaling patterns to the empirical data. Dots indicate the empirical simulation results, the solid line indicates the exact model described in Lemma~\ref{lemma:Ridge}, and the dotted line indicates the scaling curve fit to the empirical results.}
    \label{fig:stylized}
\end{figure}

    As shown via simulation in Fig.~\ref{fig:stylized}, the model introduced in this section can indeed present behaviors similar to the empirical scaling law observed in \autoref{sec:Experiments}. In this small simulation, image vectors are drawn with each component $x^{(\ell)}_i$ uniform in $[-\sqrt{3p^{-\ell}}, \sqrt{3p^{-\ell}}]$ for $\ell \in \{1, \hdots,m \}$ and $i \in \{1, \hdots, p^\ell\}$ (hence choosing $x^{(\ell)}$ such that $\E\{\bx^{(\ell)}(\bx^{(\ell)})^{\sT}\} = p^{-\ell}\id_{q^{\ell}}$). We  let $\btheta^{(\ell)} = \mathbf{1} \cdot \sqrt{r^\ell q^{-\ell}}$ for $\ell \in \{1, \hdots, m\}$, where $\mathbf{1}$ is the all ones vector of length $q^\ell$ (hence  $\btheta^{(\ell)}$ is such that $\big\Vert \theta^{(\ell)} \big\Vert^2 = r^\ell$). We use the parameters $q=2$, $p=2.1$, $r=0.99$, and $\tau=1$ and optimize over the ridge regression regularization parameter. The empirical results, represented as circles, align closely with the curves dictated by Lemma~\ref{lemma:Ridge}, represented as the solid lines. The dotted line represents the scaling law curve that is fit to the empirical data points. The fitted 
    exponents in this synthetic example, $\alpha=0.45$ and $\beta=1.02$, are similar to the classification task, and are relatively close to the exponents prescribed by \autoref{thm:main} ($\alpha=0.67$ and $\beta=1.02$).


%
\section{Discussion}
\label{sec:Discussion}

The scaling law studied in this paper admits some natural generalization.
The most obvious limitation is that we consider the model complexity as fixed, and therefore
$\Err^*$ represents the minimum test error achievable
by the model under consideration on a given task. If we vary the model complexity, as measured 
for instance by the number of parameters $m$, this quantity will change.
The common hypothesis in that case \cite{hestness2017deep,kaplan2020scaling} is
that $\Err^*(m)\approx \EBayes +C\cdot m^{-\gamma}$, where $\EBayes$ is the Bayes error, and $C,\gamma$
are parameters. This suggests to generalize the scaling law \eqref{eq:FirstScaling} as
\begin{align}
\Err(n,L) \approx \EBayes + A\cdot n^{-\alpha}+B\cdot L^{-\beta} +C\cdot m^{-\gamma}\, .\label{eq:FullScaling}
\end{align}
It would be interesting to test this extended law in future work.

We expect the coefficients $A,B$ and exponents $\alpha,\beta$ to depend on the task under consideration.
For instance, in medical image classification \cite{cai2020review}, we are often 
interested in detecting small 
patterns in a high-resolution image. In this case, the dependence on short scale components,
and hence the number of bits $L$ will be stronger. We see evidence of this pattern in the aerial image object detection task here, where the small object test error has a scaling pattern with $\beta$ an order of magnitude larger than $\alpha$.
 It would be important to develop some understand of this
variability, as much as the limits of validity of Eq.~\eqref{eq:FirstScaling}.


Moreover, the randomized strategy explored in this work provides evidence that more nuanced compression strategies may yield better results than uniformly compressing each image using the same compression settings. Additional methods of mapping images to compression levels may improve results further.

Finally, the main recommendation that emerges from this work is both simple and practical: The ``default'' approach of leaving images in lossless full resolution results in too high $L$ and too low $n$ compared to the scaling-optimal ratio. When storage is limited, even an approximate choice of the lossy compression level can result in a significantly better trade-off between quality and quantity for computer vision tasks.

\subsubsection*{Acknowledgements}
We are grateful to Joseph Gardi, Ayush Jain, Germain Kolossov, Marc Laugharn,  Rahul Ponnala,
Eren Sasoglu, and Pulkit Tandon for their feedback on this work. This work was carried
out while Andrea Montanari was on leave from Stanford and a Chief Scientist at Granica (formerly
known as Project N). The present research is unrelated to AM’s Stanford research.


\bibliographystyle{ACM-Reference-Format}
\balance
\bibliography{references}


\appendix

\section{Training Details} \label{appendix:training}
\subsection{Image Classification}
For the classification task, we take subsets of the data of 
size $n\in \{12120, 24240, 36360, 48480, 60600\}$, 
stratified by training labels, to generate different values of $n$. For each pair $(n,L)$, we train 5 different models with different random seeds, and the reported performance is the average performance across the 5 models. Each model is trained using cross entropy loss for 10 epochs with a batch size of 64 and a learning rate of 0.0001. 

\subsection{Semantic Segmentation}
In the segmentation task, we use sample sizes $n\in\{ 595, 1190, 1785, 2380, 2975\}$. Each model was trained using cross entropy loss for 80,000 steps with a batch size of 3. The learning rate has a linear warm up over the first 1,500 steps, reaching a maximum of $6 \times 10^{-6}$, then decreasing linearly for the remainder of the steps.

\subsection{Object Detection}
For the iSAID dataset, we consider $n \in \{5605, 11211,16817, 22423, 28029\}$. We train each object detection model for 12 epochs. The learning rate begins with a warm up phase, starting $2 \times 10^{-5}$ and increasing to 0.02 over the course of 500 iterations. Then, a multi-step learning rate schedule is used, multiplying the learning rate by a factor of 0.1 after the 8th and 11th epochs. Each point is averaged over two iterations from two different random seeds.

\section{Scaling Curves in Additional Metrics}\label{appendix:additional-metrics}
\begin{figure*}
    \centering
    \begin{subfigure}[t]{0.32\textwidth}
        \centering
        \setcounter{subfigure}{0}
        \includegraphics[width=\textwidth]{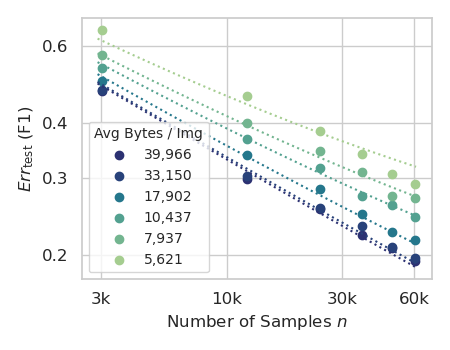}
        \caption{Classification F1 scaling in $n$}
        \label{fig:food101-f1-scaling-n}
    \end{subfigure}
    \begin{subfigure}[t]{0.32\textwidth}
        \centering
        \setcounter{subfigure}{2}
        \includegraphics[width=\textwidth]{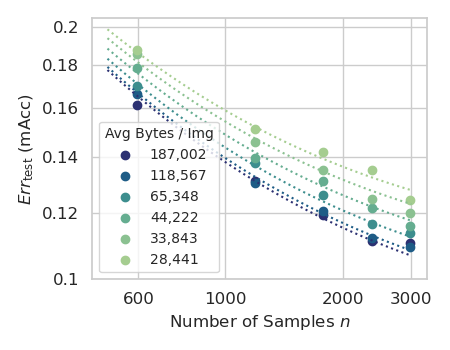}
        \caption{Segmentation mean accuracy scaling in $n$}
        \label{fig:cityscapes-macc-scaling-n}
    \end{subfigure}
    \begin{subfigure}[t]{0.32\textwidth}
        \centering
        \setcounter{subfigure}{4}
        \includegraphics[width=\textwidth]{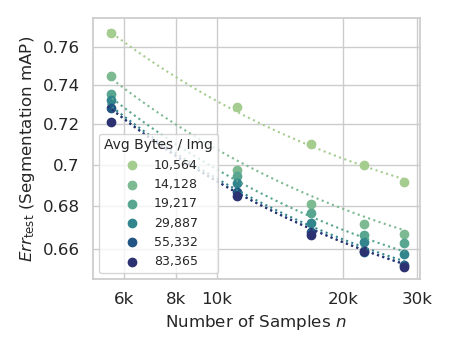}
        \caption{Mask R-CNN segmentation mean average precision scaling in $n$}
        \label{fig:isaid-segm-scaling-n}
    \end{subfigure}
    
    \begin{subfigure}[t]{0.32\textwidth}
        \centering
        \setcounter{subfigure}{1}
        \includegraphics[width=\textwidth]{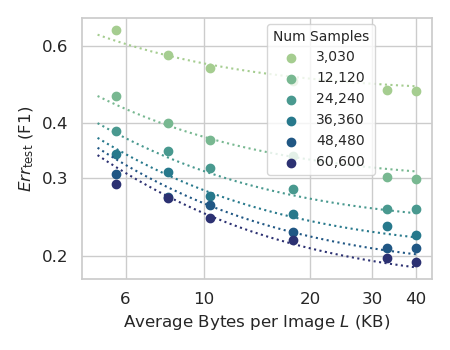}
        \caption{Classification F1 scaling in $L$}
        \label{fig:food101-f1-scaling-l}
    \end{subfigure}
    \begin{subfigure}[t]{0.32\textwidth}
        \centering
        \setcounter{subfigure}{3}
        \includegraphics[width=\textwidth]{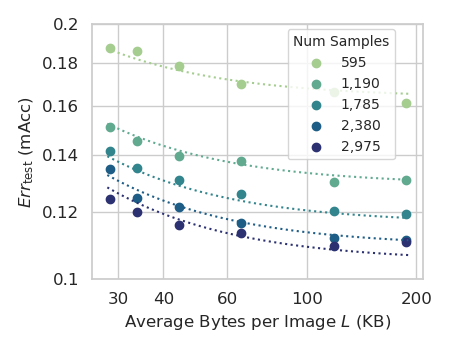}
        \caption{Segmentation mean accuracy scaling in $L$}
        \label{fig:cityscapes-macc-scaling-l}
    \end{subfigure}
    \begin{subfigure}[t]{0.32\textwidth}
        \centering
        \setcounter{subfigure}{5}
        \includegraphics[width=\textwidth]{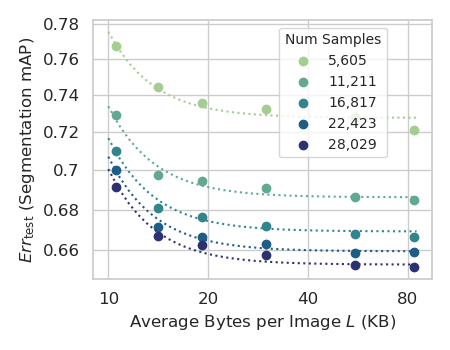}
        \caption{Mask R-CNN segmentation mean average precision scaling in $L$}
        \label{fig:isaid-segm-scaling-l}
    \end{subfigure}
    \caption{Scaling patterns for alternative metrics for each task. For the image classification task, we show multi-class F1 score as the error metric, for the semantic segmentation task, we use mean accuracy, and for the object detection task, we use the mean average precision for all objects from the Mask R-CNN's segmentation predictions.}
    \label{fig:alt-metric-scaling}
\end{figure*}

\autoref{fig:alt-metric-scaling} illustrates scaling curves with alternative error metrics. \autoref{fig:food101-f1-scaling-n}-\subref{fig:food101-f1-scaling-l} show the multi-class F1 score scaling curve for the classification task. \autoref{fig:cityscapes-macc-scaling-n}-\subref{fig:cityscapes-macc-scaling-l} shows the mean accuracy scaling curve for the segmentation task. The architecture used for the object detection task, a Mask R-CNN, can also provide instance segmentation predictions, and \autoref{fig:isaid-segm-scaling-n}-\subref{fig:isaid-segm-scaling-l} shows the mean average precision test error for these segmentation predictions. The results suggest that scaling shows a similar pattern regardless of the specific metric used to evaluate model performance.

\section{Omitted proofs}
\label{app:Proofs}
In this appendix we prove the statements of Section \ref{sec:Model}.
We begin by formally stating our assumption about the feature vectors $\bx_i$.
\begin{assumption}\label{ass:X-assumption}
Let $\bx = (\bx^{(0)},\bx^{(1)},\bx^{(2)},\dots)$ be distributed as any of the 
feature vectors $\bx_i$ and define $\bz^{(\ell)}=\bx^{(\ell)}/\sqrt{s_{\ell}}$.
In particular $\E\bz^{(\ell)} = \bfzero$ and $\E\{\bz^{(\ell)} (\bz^{(\ell)})^{\sT}\}= \id_{q^{\ell}}$.
We assume either one of the following to hold:
\begin{enumerate}
    \item[$(a)$] For each $\ell$, the entries of $\bz^{(\ell)}$ are independent sub-Gaussian.
    \item[$(b)$] For each $\ell$, $\bz^{(\ell)}$ satisfies a log-Sobolev inequality with $\ell$-independent constant $C$.
    \item[$(c)$] For each $\ell$, $\bz^{(\ell)}$ satisfies the following concentration property,
    with $\ell$-independent constant $C$. For any convex 1-Lipschitz function $\varphi$ and any $t$,
    \begin{align}
    \P\big(\varphi(\bz^{(\ell)})-\E \varphi(\bz^{(\ell)})\ge t\big)\le e^{-t^2/2C}\, .
    \end{align}
\end{enumerate}
\end{assumption}
We note in passing that condition $(b)$ above is a special case of $(c)$ and we state it
for the sake of readability.

Throughout, we write $F(n) \asymp G(n)$ of there exist constants $0<C_1<C_2$ such that
$C_1\cdot G(n)\le F(n) \le C_2\cdot G(n)$. We also use $C, C_1,\dots$ for generic constants that can change from line to line.
Throughout the proof, we assume $n\le c\, L$. The case $c\, L\le n\le c_* L^{1+2\kappa}$ is treated analogously.

\subsection{Evaluation of $\sB(m,n)$, $\sV(m,n)$, $\lambda_*$}
\label{sec:Evaluation}
We begin by evaluating the quantities appearing in Lemma \ref{lemma:Ridge}.
Begin by Eq.~\eqref{eq:lambdastar}, which we rewrite as
\begin{align}
    n = \frac{\lambda}{\lambda_*}+F(\lambda_*)\, .\label{eq:lambdastarBis}
\end{align}
Letting $\ell_*$ be such that $p^{-\ell_*-1} \le \lambda_*<p^{-\ell_*}$, we have
\begin{align*}
F(\lambda_*) &= \sum_{\ell\le \ell_*} \frac{(q/p)^{\ell}}{p^{-\ell}+\lambda_*}
+\sum_{\ell=\ell_*+1}^{m} \frac{(q/p)^{\ell}}{p^{-\ell}+\lambda_*}\\
&=\sum_{\ell\le \ell_*} \frac{(q/p)^{\ell}}{p^{-\ell}}
+\sum_{\ell=\ell_*+1}^{m} \frac{(q/p)^{\ell}}{p^{-\ell_*}}\\
&\asymp q^{\ell_*\wedge m} \asymp \lambda_*^{-1/\kappa}\wedge q^m\, .
\end{align*}
Therefore, Eq.~\eqref{eq:lambdastarBis} reduces to

\begin{align}
  \frac{\lambda}{\lambda_*}+ C_1\lambda_*^{-1/\kappa}\wedge q^m  \le n \le 
  \frac{\lambda}{\lambda_*}+ C_2\lambda_*^{-1/\kappa}\wedge q^m \, .
\end{align}
We therefore distinguish two cases (with $c_0$ a small constant:
\begin{align}
    n\le c_0 q^m&\;\; \Rightarrow \;\;  \lambda_* \asymp \frac{\lambda}{n} \vee n^{-\kappa}
    \, ,\\
     n> c_0 q^m&\;\; \Rightarrow \;\;  \lambda_* \asymp \frac{\lambda}{n}\,  .
\end{align}
Pursuing the first case, we have
\begin{align*}
    \sD(m) & \asymp \sum_{\ell=0}^{m}\frac{q^{\ell} p^{-2\ell}}{(\lambda_*+p^{-\ell})^2}\\
    &\asymp  \sum_{\ell=0}^{\ell_*} q^{\ell}+\sum_{\ell=\ell_*+1}^{m} q^{\ell}
    p^{-2(\ell-\ell_*)}\\
    &\asymp q^{\ell_*\wedge m}\, .
\end{align*}
Further, by comparing the terms $\ell=\ell_*$ of $\sD(m)$ and $F(\lambda_*)$
we obtain that $\sD(m)\le (1-c_1) F(\lambda_*)$ for some positive constant $c_1$.

Substituting in the formula for $\sV$, we get 
\begin{align}
    \sV(m,n) \asymp \frac{1}{n}q^{\ell_*\wedge m} \asymp \big(\lambda^{-1/\kappa} n^{-1+1/\kappa}
    \wedge 1\big)\, .
\end{align}

We further have (using $rp<1$)
\begin{align*}
  \sum_{\ell=0}^{m}
\frac{s_{\ell}\|\btheta^{(\ell)}\|^{2}} {(s_{\ell}+\lambda_*)^2} &\asymp 
\sum_{\ell=0}^{m}\frac{r^{\ell}p^{-\ell}} {(p^{-\ell}+\lambda_*)^2} \\
& \asymp\sum_{\ell=0}^{\ell_*}\frac{r^{\ell}} {p^{-\ell}} 
+\sum_{\ell=\ell_*+1}^{m}\frac{r^{\ell}p^{-\ell}} {p^{-2\ell_*}}\\
& \asymp 1\, ,
\end{align*}
whence 
\begin{align*}
 \sB(m,n) \asymp  \lambda_*^2 \asymp \Big(\frac{\lambda}{n}\Big)^2 \vee n^{-2\kappa}\, .
\end{align*}

\subsection{Proof of Lemma \ref{lemma:Ridge}}

This lemma is an application of Proposition 4.3 in \cite{cheng2022dimension}.
We only need to check the assumptions of that proposition, and compute error terms, which we do below
(following notations from \cite{cheng2022dimension}, end denoting by $b_{\ell}:=(1-q^{\ell})/(1-q)$):
\begin{enumerate}
\item \emph{Effective dimension $\sd_{\bSigma}$.} Note that, for $k\in [b_{\ell}+1,b_{\ell+1}]$, we
have
\begin{align}
    \sum_{i\ge k} \sigma_i \le C\sum_{j\ge \ell} p^{-j} q^{j}\le Cq^{\ell} \sigma_k\,.
\end{align}
and for $k\le n$, the last expression is at most $Cn\sigma_k$. Therefore $\sd_{\bSigma} \asymp n$.
\item  \emph{Effective regularization $\lambda_*(\lambda)$.} This quantity is analyzed in the previous section, implying that for $n \le c_0L$, $\lambda_*(\lambda) \asymp (n^{-\kappa}\wedge(\lambda/n))$.
In particular, $\lambda\in [n\lambda_*(\lambda)/M,n\lambda_*(\lambda)M]$
under the assumptions of the lemma. 
(Note that Proposition 4.3 in \cite{cheng2022dimension} states this condition as
$\lambda\in [n\lambda_*(0)/M,n\lambda_*(0)M]$, but the condition checked here is also sufficient.)
\item \emph{Coefficient $\rho(\lambda)$.} Recalling the general definition from \cite{cheng2022dimension}, we have:
\begin{align}
   \rho(\lambda) = \frac{\sum_{\ell=0}^m\|\btheta^{(\ell)}\|^2/(s_{\ell}+\lambda_*)} {F(\lambda_*)\sum_{\ell=0}^m\|\btheta^{(\ell)}\|^2/s_{\ell}} \, ,
\end{align}
where $\lambda_*$ is determined by Eq.~\eqref{eq:lambdastar} (we note in passing a difference in notation with respect to \cite{cheng2022dimension}, since we use $\btheta$
for what is denoted by $\bbeta$ there.)

Various terms are estimated analogously to  the last section.
First consider the denominator
\begin{align*}
F(\lambda_*) &\asymp \lambda_*^{-1/\kappa} \asymp \Big(\frac{\lambda}{n}\Big)^{-1/\kappa}
\wedge n\, ,\\
\sum_{\ell=0}^m\|\btheta^{(\ell)}\|^2/s_{\ell} & \asymp
\sum_{\ell=0}^m (rp)^{\ell}\asymp 1\, .
\end{align*}
Considering next the numerator:
\begin{align*}
\sum_{\ell=0}^m\frac{\|\btheta^{(\ell)}\|^2}{s_{\ell}+\lambda_*}
& \asymp \sum_{\ell=0}^m\frac{r^{\ell}}{p^{-\ell}+\lambda_*}\\
&  \asymp \sum_{\ell=0}^{\ell_*}\frac{r^{\ell}}{p^{-\ell}}
+ \sum_{\ell=\ell_*+1}^{m}\frac{r^{\ell}}{p^{-\ell_*}}\asymp 1
\end{align*}
Putting everything together, we obtained
\begin{align}
   \rho(\lambda) =  \Big(\frac{\lambda}{n}\Big)^{1/\kappa}
\vee \frac{1}{n}\, .
\end{align}
\end{enumerate}

We finally applying Proposition 4.3 in \cite{cheng2022dimension}.
to the second term of \eqref{eq:BasicDec}, which reduces to the setting
of  \cite{cheng2022dimension} with $\tau^2$ replaced by $\tilde\tau^2_m$.
We obtain that,
for $\lambda\in [n^{-\kappa}/M,n^{-\kappa}M]$, with $M$ an arbitrary constant,
and dropping the arguments from $\cuB, \cuV,\sB,\sV$,
\begin{align}
  |\cuV-\sV|&\le O\big(n^{-1+\eps}\big)\cdot\sV\, ,\\
    |\cuB-\sB|&\le O\big(\rho(\lambda)^{-1/2}n^{-1+\eps}\big)\cdot\sB = 
     O\big(n^{-1/2+\eps}\big)\cdot \sB\, ,
\end{align}
which proves our claim.

\subsection{Proof of Theorem \ref{thm:main}}

Collecting the expressions for bias and variance from  Section \ref{sec:Evaluation}
\begin{align*}
 K_V\big(\lambda^{-1/\kappa} n^{-1+1/\kappa}\wedge 1)  \le   \sV(m,n) &\le K_V'\big(\lambda^{-1/\kappa} n^{-1+1/\kappa}
    \wedge 1\big)\, ,\\
K_B  \Big[\Big(\frac{\lambda}{n}\Big)^2 \vee n^{-2\kappa}\Big] \le \sB(m,n) & \le K_B'   
\Big[\Big(\frac{\lambda}{n}\Big)^2 \vee n^{-2\kappa}\Big]\, .
\end{align*}
Using the choice $\lambda \asymp n^{(\kappa+1)/(2\kappa+1)}$ as in the statement on the theorem,
we get (eventually adjusting the constants):
\begin{align*}
 K_V n^{-\alpha} \le   \sV(m,n) &\le K_V'n^{-\alpha}\, ,\\
K_B n^{-\alpha}  \le \sB(m,n) & \le K_B'   n^{-\alpha} \, .
\end{align*}
Substituting in Lemma  \ref{lemma:Ridge}, we get
\begin{align*}
\E_{\eps}\Err(n,L) \le \tau^2 + A\cdot n^{-\alpha}+
B\cdot (r/p)^{m}\, .
\end{align*}
The upper bound in the theorem follows by recalling the relation between $m$ and $L$.
The lower bound follows by modifying the last two steps.
\end{document}